\documentclass{article}
\usepackage[numbers]{natbib}

\usepackage[english]{babel}

\usepackage[letterpaper,top=2cm,bottom=2cm,left=3cm,right=3cm,marginparwidth=1.75cm]{geometry}

\usepackage{amsmath}
\usepackage{graphicx}
\usepackage[colorlinks=true, allcolors=blue]{hyperref}
\usepackage{threeparttable}
\title{Trainable Adaptive Activation Function Structure (TAAFS)  Enhances Neural Network Force Field Performance with Only Dozens of Additional Parameters}
\author{Enji Li\\
 \small State Key Lab of Processors, Institute of Computing Technology, Chinese Academy of Sciences \\
\small University of Chinese Academy of Sciences \\
     lienji23s\@ ict.ac.cn}

\begin{document}
\maketitle 

\begin{abstract}
At the heart of neural network force fields (NNFFs) is the architecture of neural networks, where the capacity to model complex interactions is typically enhanced through widening or deepening multilayer perceptrons (MLPs) or by increasing layers of graph neural networks (GNNs). These enhancements, while improving the model's performance, often come at the cost of a substantial increase in the number of parameters. By applying the Trainable Adaptive Activation Function Structure (TAAFS), we introduce a method that selects distinct mathematical formulations for non-linear activations, thereby increasing the precision of NNFFs with an insignificant addition to the parameter count. In this study, we integrate TAAFS into a variety of neural network models, resulting in observed accuracy improvements, and further validate these enhancements through molecular dynamics (MD) simulations using DeepMD.
\end{abstract}


\section{Introduction}
Molecular dynamics (MD) simulations are a critical research tool widely applied across various disciplines, including chemical engineering, biomedicine, materials science, and physics. MD models encompass several methodologies, such as ab initio molecular dynamics (AIMD), potential function methods, and neural network-based force fields, which rely on mathematical models.

The advent of artificial intelligence and the computation of high-precision ab initio data have led to the development of neural network force fields. These methods offer near-ab-initio accuracy while maintaining computational costs similar to those of potential function methods, making them highly attractive for applications in computational materials science and molecular dynamics simulations. Prominent machine learning force field models include SNAP~\cite{THOMPSON2015316}, SIMPLE-NN~\cite{lee2019simple}, HDNNP~\cite{Behler_2014}, DeePMD~\cite{wang2018deepmd}, NEP~\cite{fan2022gpumd}, BPNN~\cite{dai1997effects}, ANI~\cite{smith2017ani}, chgnet~\cite{deng_2023_chgnet}, MACE~\cite{Batatia2022mace}, among others.

At the heart of neural network force fields lies the neural network itself. Enhancing the expressive power of neural networks has been an ongoing area of exploration. In multi-layer perceptrons (MLPs~\cite{pinkus1999approximation}), this often involves increasing the width and depth of the network. However, the increase in parameters does not proportionally enhance expressive power, and performance can plateau once a certain parameter threshold is reached. Graph neural networks (GNNs) possess unique properties that make them suitable for machine learning force fields, but they also come with substantial parameter counts. For instance, CHGNet has 413K parameters, MACE has 4.69M, whereas DeePMD typically has only 26K. Increasing GNN layers leads to significant increases in parameter count.

Recently, the KAN~\cite{liu2024kan} introduced trainable activation functions on edges, with nodes responsible only for summation. This approach reduced parameter complexity from O(N²) to O((G+K)N²). Despite its innovative concept, the practical application of KAN did not meet expectations; replacing MLPs with KAN in DP resulted in no performance improvement, with parameter count and training time increasing by approximately 8 times and 10 times, respectively.

Although KAN did not achieve expected results in my tasks, its method of training activation functions offered significant inspiration. Drawing upon the KAN framework, I propose the Trainable Adaptive Activation Function Structure (TAAFS). Unlike other trainable activation functions based on fixed transformations and combinations, TAAFS allows for the free selection of mathematical formulas for nonlinear computations. The shape of the activation function curve adapts dynamically based on data and selected mathematical formulas. The primary workflow includes normalization, mathematical formula computation, and training coefficients, detailed in Chapter 3.

Ultimately, applying TAAFS to models like DP, ANI2, and chgnet resulted in over 10\% accuracy improvements with only a marginal increase in parameters—ranging from tens to hundreds, less than 0.2\% of the total parameter count. Experimental results are presented in Chapter 5.

The main contributions of this work are as follows:
\begin{enumerate}
\item Transfer of KAN's trainable activation function to MLPs: Within this structure, we can arbitrarily select mathematical formulas as activation bases.
\item First Implementation of Trainable Activation Functions in Neural Network Force Fields: To our knowledge, this is the first time such an approach has been employed in neural network force fields.
\item Enhancement of Model Accuracy Using Mathematical Formulas in DP Networks: We improved model accuracy using multiple mathematical formulas for fitting and validated the results using LAMMPS~\cite{thompson2022lammps} for MD simulations.
\item Successful Application of Trainable Activation Functions in GNN Layers: Achieving performance gains by incorporating trainable activation functions into GNN layers.
\end{enumerate}

\section{Related Work}

In the absence of activation functions, no matter how many layers a neural network possesses, it can only function as a linear model. This limitation arises because the composition of multiple linear transformations can always be reduced to a single linear transformation. The introduction of activation functions infuses non-linearity into the network, thereby empowering it to learn and approximate highly complex and non-linear mappings that are essential for capturing intricate patterns within data. Consequently, activation functions are indispensable components in the architecture of artificial neural networks.

Activation functions can be broadly classified into two categories: fixed activation functions, which have predefined forms and parameters that do not change during training; and trainable activation functions, which possess parameters that can be optimized during the training process to better fit the data. The adaptability of trainable activation functions allows for greater flexibility and potentially improved performance in specific tasks.

Below, I will provide a brief introduction to the two categories of activation functions.

\subsection{Fixed-shape activation function}
Fixed activation functions include traditional ones such as Sigmoid~\cite{university1988continuous} and Hyperbolic Tangent (tanh)~\cite{xiao2005simple}, as well as their variants like SiLU (Sigmoid Linear Unit)~\cite{dahl2013improving}, ReLU (Rectified Linear Unit)~\cite{glorot2011deep}, and Leaky ReLU (LReLU)~\cite{maas2013rectifier}.

\begin{itemize}
    \item \textbf{Sigmoid}:
        Once popular for mapping predictions to probabilities, sigmoid functions are now less favored due to the vanishing gradient problem in deep networks.
 
        \begin{equation}
        \sigma(x) = \frac{1}{1 + e^{-x}}
        \end{equation}

    \item \textbf{Hyperbolic Tangent (tanh)}:
        Similar to the sigmoid function but maps inputs to the range (-1, 1), which can sometimes lead to better convergence.
        \begin{equation}
        \tanh(x) = \frac{e^x - e^{-x}}{e^x + e^{-x}}
        \end{equation}
\end{itemize}


\begin{itemize}
    \item \textbf{ReLU (Rectified Linear Unit)}:
        Introduces non-linearity while mitigating the vanishing gradient problem, though it can suffer from "dying ReLU" issues.
        \begin{equation}
        \text{ReLU}(x) = \max(0, x)
        \end{equation}

    \item \textbf{Leaky ReLU (LReLU)}:
        Addresses the "dying ReLU" problem by allowing a small, non-zero gradient when the unit is not active.
        \begin{equation}
        \text{LReLU}(x) = \max(\alpha x, x), \quad \text{where } \alpha \text{ is a small constant}
        \end{equation}

    \item \textbf{SiLU (Sigmoid Linear Unit)}:
        Combines the benefits of sigmoid and ReLU, providing smooth gradients and avoiding dead neurons.
        \begin{equation}
        \text{SiLU}(x) = x \cdot \sigma(x) = \frac{x}{1 + e^{-x}}
        \end{equation}
\end{itemize}


\subsection{Trainable activation functions}
Trainable activation functions most commonly include parameterized versions of standard activation functions, as well as combinations of multiple basic activation functions to create more complex and adaptive activations. Below are some examples.

 
\textbf{Parametric Exponential Linear Unit (PELU)}: Introduces non-linearity with two trainable parameters 
$\gamma$ and 
$\beta$, which respectively adjust the saturation level and slope of the activation function. This circumvents the need for manual setting of parameters as seen in ELU, allowing PELU to adapt dynamically during training via backpropagation~\cite{trottier2017parametric}.

\begin{equation}
\text{PELU}(x) =
\begin{cases}
\beta \cdot \frac{x}{\gamma}, & \text{if } x \geq 0 \\
\beta \cdot \left(\exp\left(\frac{x}{\gamma}\right) - 1\right), & \text{otherwise}
\end{cases}
\end{equation}

where $\beta$ and $\gamma$ are trainable parameters.

\textbf{Parametric ReLU (PReLU)}: Extends the standard ReLU~\cite{he2015delving} by introducing a learnable parameter $\alpha$ that modifies the response to negative inputs. This allows the activation function to partially learn its form from the training data, mitigating the "dying ReLU" problem where neurons can become inactive and only output zero.

\begin{equation}
\text{PReLU}(x) =
\begin{cases}
x, & \text{if } x > 0 \\
\alpha \cdot x, & \text{otherwise}
\end{cases}
\end{equation}

where $\alpha$ is learned concurrently with the model parameters using gradient-based methods.


\textbf{Exponential Linear Sigmoid Squashing (EliSH)}: Combines properties of both exponential linear units (ELUs) and sigmoid functions. It introduces non-linearity while maintaining a smooth gradient flow, even for large input values. EliSH~\cite{basirat2018quest} is designed to provide a balance between computational efficiency and performance, making it suitable for various neural network architectures.

The EliSH function is defined as follows:
\begin{equation}
\text{EliSH}(x) =
\begin{cases}
\frac{x}{1 + e^{-x}}, & \text{if } x \geq 0 \\
\frac{e^x - 1}{1 + e^{-x}}, & \text{otherwise}
\end{cases}
\end{equation}

In this formulation, the negative part of EliSH results from the multiplication of the ELU and sigmoid functions, whereas the positive part shares similarities with the Swish function.
HardELiSH, meanwhile, is constructed by multiplying the HardSigmoid and ELU functions in the negative domain and combining the HardSigmoid and linear functions in the positive domain.

\textbf{Mexican Hat Linear Unit (MeLU)}: The Mexican Hat Linear Unit ~\cite{maguolo2021ensemble} addresses the issues of unstable learning associated with trainable parameters. Unstable learning can lead to decreased accuracy and increased generalization error when model performance varies significantly in response to slight changes in data or parameters. Mexican hat-type functions offer a smoother curve compared to ReLU, which helps prevent saturation and allows for optimal performance.

The core function $ f $ is defined as:
\begin{equation}
f_{\gamma,\lambda}(x) = \max(\lambda - |x - \gamma|, 0),
\end{equation}
where $\lambda$ and $\gamma$ are real numbers. This function returns zero when $ |x - \gamma| > \lambda $. It increases with a derivative of $+1$ between $\gamma - \lambda$ and $\gamma$, then decreases with a derivative of $-1$ between $\gamma$ and $\gamma + \lambda$.

The MeLU is defined for each layer as:
\begin{equation}
\text{MeLU}(x) = \text{PReLU}(x) + \sum_{j=1}^{k-1} c_j f_{\gamma_j, \lambda_j}(x),
\end{equation}
where $ k $ represents the number of learnable parameters per neuron. The $ c_j $ parameters are learnable real numbers, while $\gamma_j$ and $\lambda_j$ are fixed parameters.

\begin{figure}[t!]
  \centering
  \includegraphics[scale=0.18]{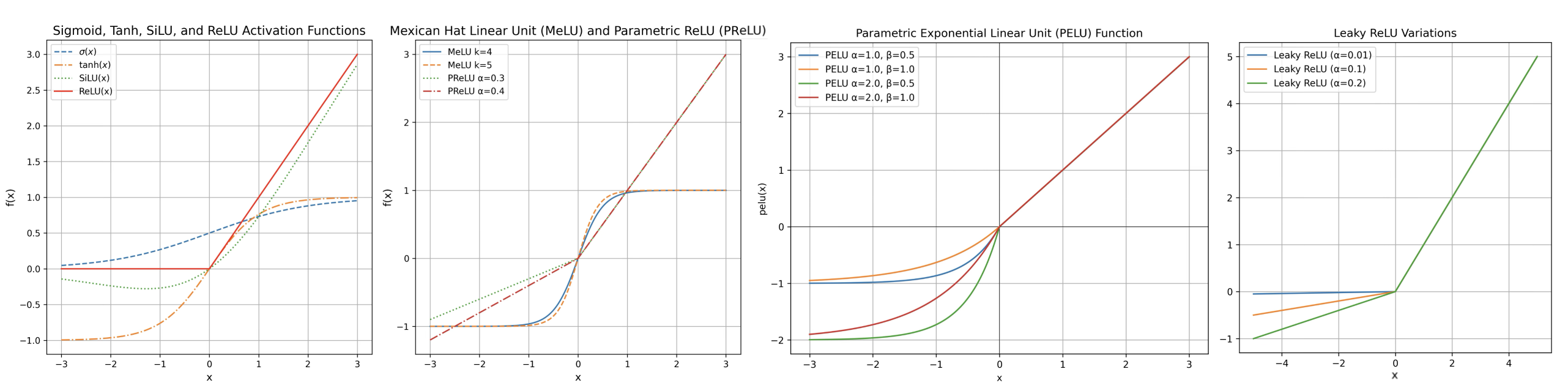}
  \caption{
Activation Function Curves
  }
  \label{fig:1}
\end{figure}

In this study, the neural network force field models selected all employ fixed activation functions. Therefore, the experimental focus is on demonstrating the feasibility of TAAFS (Trainable Activation and Adaptive Force Field Scheme) and its performance enhancement relative to fixed activation functions, rather than comparing it with trainable activation functions. Fig\ref{fig:1} presents graphical representations of several activation functions, which can be compared against the activation function curves derived from TAAFS, providing an intuitive visualization of their differences.

\section{Method }
We adopt the approach proposed by KAN, utilizing a Trainable Adaptive Activation Function Structure (TAAFS) within neural networks. The following sections outline the working principle of TAAFS, the mathematical formulaused, and comparisons with traditional activation functions.

\subsection{Working Principle}

The core idea behind TAAFS is to replace fixed activation functions with adaptive ones that can be trained alongside the network parameters. By fitting mathematical formulato the data, these activation functions can dynamically adjust their shape during training to better match the underlying patterns in the dataset. This adaptability allows the model to learn more complex and nuanced representations compared to using static activation functions.

Traditional Fixed Activation Function Forward Propagation Steps :
\begin{enumerate}
    \item Weighted Summation: $\mathbf{z} = \mathbf{Wx} $, where $\mathbf{W}$ is the weight matrix, $\mathbf{x}$ is the input vector.
    \item Nonlinear activation is performed using activation functions, such as the Hyperbolic Tangent (Tanh) and the Sigmoid Linear Unit (SiLU):$a(\mathbf{z})$.
    \item Addition of Bias : $\mathbf{y} = a(\mathbf{z}) + \mathbf{b}'$ if necessary, where $\mathbf{b}'$ is another bias vector that may be added after the activation.
\end{enumerate}
During backpropagation, the weights $\mathbf{W}$ and biases $\mathbf{b}$ are adjusted.
\\

Trainable Adaptive Activation Function Structure Forward Propagation Steps :
\begin{enumerate}
    \item Weighted Summation: $\mathbf{z} = \mathbf{Wx} $.
    \item Normalization can be performed using activation functions such as the Hyperbolic Tangent (Tanh), or alternatively, through methods like the difference quotient:$\mathbf{a} = \mathbf{f(z)} $. 
    \item Setting up a grid for mathematical formulaor determining the number of iterations $N$: This step depends on the mathematical formulaor type used.
    \item Mathematical Formulaor Evaluation: $p(\mathbf{a}; \theta)$, where $p(\cdot; \theta)$ represents the mathematical formulaor with trainable parameters $\theta$.
    \item Addition of Bias: $\mathbf{y} = p(\mathbf{a}; \theta) + \mathbf{b}$, where $\mathbf{b}$ is the bias vector added after polynomial evaluation.
\end{enumerate}
During backpropagation, the weights $\mathbf{W}$, biases $\mathbf{b}$, and polynomial coefficients $\theta$ are adjusted.
\\

\subsection{Mathematical Formula}
We employ various mathematical formulato fit the data, including B-spline basis function ( Bline ), Fourier, Gaussian Radial Basis Function (GRBF), Legendre, Hermite, Chebyshev of the second kind (Chebyshev2), Chebyshev of the first kind (Chebyshev1), Bessel, and Jacobi. These mathematical formulahave been extensively used in mathematics, physics, statistics, and other fields. One class of these mathematical formularequires a grid setup for evaluation, such as Bline, Fourier, and GRBF. The other class only considers the number of iterations for computation, like Chebyshev polynomials of the first kind (Chebyshev1) and Hermite polynomials.The following is an introduction to a few representative multiples.

\textbf{B-spline functions} are piecewise polynomial functions that are widely used for constructing smooth curves and surfaces. Given a knot vector $ U = \{u_0, u_1, \ldots, u_m\} $ and a degree $ p $, the B-spline basis function $ N_{i,p}(u) $ is defined recursively using the Cox-de Boor formula:

For $ p=0 $:
\begin{equation}
N_{i,0}(u) = 
  \begin{cases} 
   1 & \text{if } u_i \leq u < u_{i+1} \\
   0 & \text{otherwise}
  \end{cases}
\end{equation}

For $ p > 0 $:
\begin{equation}
N_{i,p}(u) = \frac{u - u_i}{u_{i+p} - u_i} N_{i,p-1}(u) + \frac{u_{i+p+1} - u}{u_{i+p+1} - u_{i+1}} N_{i+1,p-1}(u)
\end{equation}

The advantages of B-splines include local support, where each B-spline function affects only a limited interval, and smoothness, ensuring continuous derivatives up to order $ p-1 $. However, the recursive nature can make implementation more complex.

\textbf{The Fourier series represents periodic functions} as a sum of sine and cosine terms. For a function $ f(x) $ with period $ 2\pi $, the Fourier series is given by:
\begin{equation}
f(x) = \frac{a_0}{2} + \sum_{n=1}^{\infty} \left( a_n \cos(nx) + b_n \sin(nx) \right)
\end{equation}
where the coefficients $ a_n $ and $ b_n $ are calculated as follows:
\begin{equation}
a_n = \frac{1}{\pi} \int_{-\pi}^{\pi} f(x) \cos(nx) \, dx, \quad b_n = \frac{1}{\pi} \int_{-\pi}^{\pi} f(x) \sin(nx) \, dx
\end{equation}

Advantages of Fourier series include their excellent representation of periodic data and the orthogonality of sine and cosine functions. However, they suffer from the Gibbs phenomenon, which causes oscillations near discontinuities in the function.

\textbf{First-kind Chebyshev} polynomials $ T_n(x) $ are defined on the interval $[-1, 1]$ and have the property of minimizing the maximum deviation from zero. They are given by the recurrence relation:
\begin{equation}
T_0(x) = 1, \quad T_1(x) = x, \quad T_{n+1}(x) = 2xT_n(x) - T_{n-1}(x)
\end{equation}
Alternatively, they can be expressed using trigonometric identities:
\begin{equation}
T_n(x) = \cos(n \arccos(x))
\end{equation}

Advantages of first-kind Chebyshev polynomials include their minimax property, which minimizes the maximum error in polynomial approximation, and efficient computation through recurrence relations. However, they are defined on the interval $[-1, 1]$, requiring appropriate scaling for other ranges.

Admittedly, the repertoire of applicable mathematical formulas is not limited to those enumerated herein. A vast array of alternative mathematical formulations, including but not limited to rational functions and wavelet transforms, have been explored. Due to constraints imposed by table width and less favorable convergence outcomes, the results pertaining to these methods are omitted from the experimental presentation.

\subsection{Contrast}
Advantages of Using Trainable Adaptive Activation Function Structure:

1 : Wider Output Range: Although all activation functions typically involve normalization, polynomial fitting allows for a broader and more flexible expression range. This flexibility can better capture complex patterns in the data.

2 : Freedom from Traditional Shapes: Polynomial fitting enables the activation function to break free from the conventional shapes imposed by standard activation functions, providing greater adaptability to different datasets and tasks.

3 : Improved Smoothness: Polynomial fitting ensures that the resulting activation curve is smoother, which can contribute to more stable training processes and potentially better generalization performance.

4 : Minimal Increase in Parameters: The approach requires only a modest increase in parameters, making it relatively simple to implement. Users need only specify the polynomial degree and the number of grid points or iterations.

Disadvantages of Using TAAFS:

1 : Increased Training Time Due to Serial Operations: Training the parameters of the polynomial-based activation function is a serial operation, which can add to the overall training time compared to using fixed activation functions.

2 : Sensitivity to Fitting Method Selection: The choice of fitting method can significantly influence the training outcomes. Selecting an inappropriate fitting technique might lead to suboptimal performance or convergence issues.

\section{Experiments Model}
This study employs four models for experimentation: the Deep Potential model (DP), ANI2, PAINN, and ChgNet.

\begin{figure}[t!]
  \centering
  \includegraphics[scale=0.25]{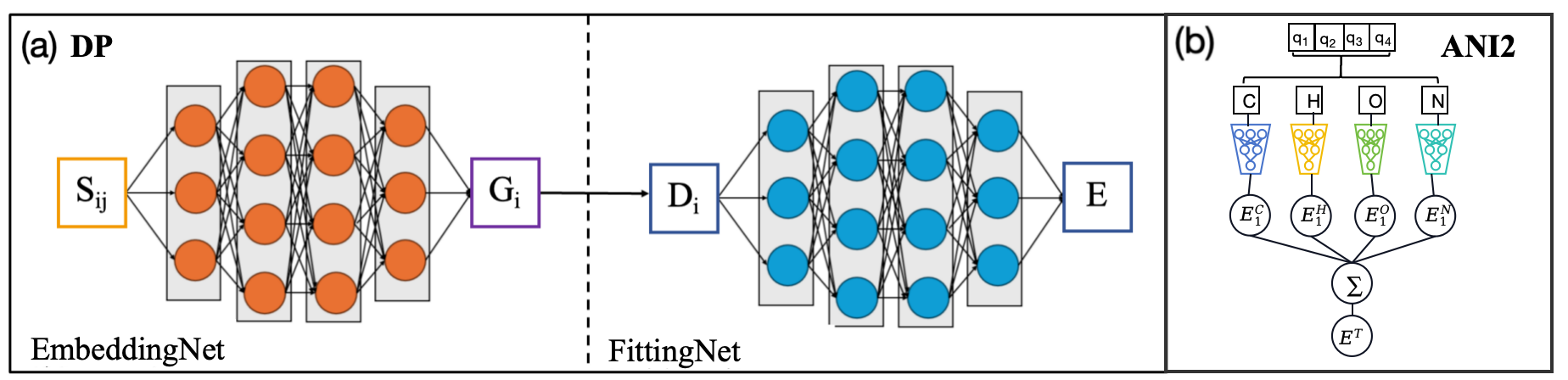}
  \caption{
Deep Potential model and ANI2
  }
  \label{fig:4-1}
\end{figure}

Deep Potential Model (DP): The DP model comprises two networks arranged sequentially: an embedding network and a fitting network, as illustrated in fig\ref{fig:4-1}(a). The embedding network transforms atomic configurations into descriptors that capture the local environment of each atom, while the fitting network maps these descriptors to atomic energies, thereby enabling the prediction of the total potential energy of the system.
In the context of Deep Potential Model (DP), two optimizers are employed: ADAM~\cite{kingma2014adam} and RLEKF~\cite{hu2023rlekf}. ADAM is a well-established optimizer, yet it tends to train relatively slowly, often necessitating several hundred epochs to converge. Conversely, RLEKF is a second-order optimizer capable of converging to the same level of precision as ADAM within just a few epochs. However, RLEKF demands significantly more memory and each epoch takes longer than with the ADAM optimizer. Despite these drawbacks, RLEKF shows a rapid decrease in RMSE, achieving lower error rates quickly.
Given this comparative analysis, the utilization of the RLEKF optimizer predominates in DPI testing due to its capability to quickly converge to a lower RMSE. Code from:\url{https://github.com/LonxunQuantum/PWMLFF}.

ANI2: This model specifically targets the elements carbon (C), hydrogen (H), oxygen (O), and nitrogen (N)~\cite{berenger2024ani}. ANI2 constructs separate neural networks for each element to compute their contributions individually. These individual contributions are then aggregated to calculate the total energy of the molecular system, as illustrated in fig\ref{fig:4-1}(b).Code from:\url{https://github.com/aiqm/torchani/tree/master}.

PAINN and ChgNet: Both PAINN and ChgNet are based on Graph Neural Network (GNN) architectures, which are designed to handle complex interactions within molecular structures. These GNNs process atoms as nodes and bonds as edges in a graph representation of molecules. They iteratively update node features by aggregating information from neighboring nodes, allowing for sophisticated modeling of interatomic forces and electronic properties. Due to their complexity and effectiveness in capturing intricate patterns in molecular data, both models represent advanced approaches in the field of computational chemistry.

\section{Experiments }
All experiments in this chapter are conducted in a progressive manner, with the results from each preceding experiment being applied to subsequent experiments.
\subsection{Experiment 1:Best practice testing}

In this experiment, we conducted tests using the RLEKF optimizer in the DP  with a smaller Ag dataset and used the built-in example dataset from the ANI2 code for testing. The DP network consists of two serially connected networks: EmbeddingNet and FittingNet. In contrast, the ANI2 model comprises parallel networks for four elements: carbon (C), hydrogen (H), oxygen (O), and nitrogen (N). The experiments were carried out according to the following schemes:

1:Global Activation Function: Using a singletrainable adaptive activation function across the entire network.

2:Per-Network Activation Functions: Using onetrainable adaptive activation function per network.

3:Per-Layer Activation Functions: Using onetrainable adaptive activation function per layer.

4:Per-Neuron Activation Functions: Using onetrainable adaptive activation function per neuron.

The fitting scheme for thetrainable adaptive activation functions was set to third-order Bline. The parameter counts for each configuration of DP model are summarized in the table \ref{tab:parameter1}.


\begin{table}[t]
  \begin{center}
    \caption{Comparison of DP Parameters Across Different Schemes}
    \label{tab:parameter1}
    \begin{threeparttable}
    \begin{tabular}{l l l l l l }
    \hline
        ~ & Original & scheme 1 & scheme 2 & scheme 3 & scheme 4  \\ \hline
        Parameter & 26550 & 26558 & 26566 & 26598 & 26775  \\ 
        Ratio & 100\% & 100.03\% & 100.06\% & 100.18\% & 100.85\% \\ \hline
    \end{tabular}
    \begin{tablenotes}    
        \item[]  DP network structure :embedding network  [25,25,25],  fitting network [50,50,50,1] .
    \end{tablenotes}         
  \end{threeparttable}
  \end{center}
\end{table}

From the results, it can be observed that in the DP network, schemes 1, 2, and 3 exhibited similar computational times. As the scope of application for eachtrainable adaptive activation function became more localized, performance improved, with the best results obtained when using a separatetrainable adaptive activation function per layer. However, using a separatetrainable adaptive activation function per neuron (scheme 4) yielded poor performance. Examination of the training logs revealed that during the first epoch, the error in energy and forces stopped decreasing after processing only a small portion of the data. This issue arose due to rapid gradient vanishing caused by the excessive number of neurons.

\begin{table}[t]
  \begin{center}
    \caption{Convergence Results of the DP Model on the Test Set}
    \label{tab:error_energy}
    \begin{threeparttable}
    \begin{tabular}{lllllll} \hline
        Scheme & E/atom, eV & E Ratio & F, eV/$ \mathring{A}$ & F Ratio & s/epoch & Time Ratio  \\  \hline
        Original  & 1.90E-04 & 100.00\% & 1.13E-02 & 100.00\% & 26.2 & 100.00\%  \\ 
        scheme 1 & 1.82E-04 & 95.75\% & 1.12E-02 & 98.80\% & 51.6 & 196.95\%  \\ 
        scheme 2 & 1.80E-04 & 94.93\% & 1.11E-02 & 98.46\% & 51.7 & 197.33\%  \\ 
        scheme 3 & 1.77E-04 & 93.24\% & 1.12E-02 & 98.74\% & 52.1 & 198.85\%  \\ 
        scheme 4 & 3.91E+03 & 90697673\% & 3.36E-01 & 2969\% & 594 & 2267\% \\  \hline
    \end{tabular}
  \end{threeparttable}
  \end{center}
\end{table}

In the ANI2 network, while the per-layer approach showed good performance, the computational time was significantly higher. Therefore, scheme 4 was not tested in ANI2. For subsequent experiments, except for ANI2 which used different networks withtrainable adaptive activation functions, all other configurations utilized a per-layertrainable adaptive activation function.
\begin{table}[t]
  \begin{center}
    \caption{Convergence Results of ANI2 on example data}
    \label{tab:error_energy}
    \begin{threeparttable}
    \begin{tabular}{llll} \hline
        Example Data & RMSE & Ratio & second/epoch  \\ \hline
        Original  & 1.894896437 & 100\% & 1s  \\ 
        scheme 1 & 2.119502814 & 111.85\% & 10s  \\ 
        scheme 2 & 1.530401059 & 80.76\% & 10s  \\ 
        scheme 3 & 1.350248646 & 71.26\% & 129s  \\ 
        scheme 4 & / & / & / \\  \hline
    \end{tabular}
  \end{threeparttable}
  \end{center}
\end{table}

\subsection{Experiment 2:Big data set testing}


Based on the results from Experiment 1, we conducted relevant tests usingtrainable adaptive activation functions on a layer-by-layer basis with larger datasets.

For ANI2, we tested using the dataset provided in the original paper. Despite some discrepancies between our implementation and the paper's results, the use oftrainable adaptive activation functions led to precision improvements of 32\% and 18\%. The training process consistently yielded good accuracy, with a particularly significant reduction in initial error.

\begin{figure}[t!]
  \centering
  \includegraphics[scale=0.38]{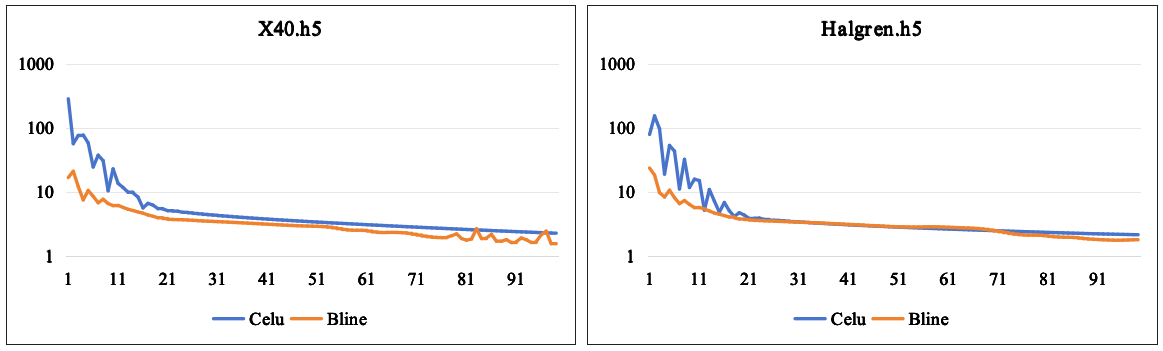}
  \caption{
Convergence Curve of ANI2
  }
  \label{fig:ANI2_1}
\end{figure}

\begin{table}[t]
  \begin{center}
    \caption{Convergence Results of ANI2}
    \label{tab:error_energy}
    \begin{threeparttable}
    \begin{tabular}{lll|lll}\hline
        X40.h5 & RMSE & Ratio & Halgren.h5 & RMSE & Ratio  \\ \hline
        Thesis results & 2.44 & / & Thesis results & 1.77 & /  \\ 
        Experimental results & 2.28 & 1 & Experimental results & 2.16 & 1  \\ 
        Bline & 1.56 & 68.42\% & Bline & 1.78 & 82.36\% \\ \hline
    \end{tabular}  
  \end{threeparttable}
  \end{center}
\end{table}

In the DP network, we used the same dataset as in paper (Training one DeePMD Model in Minutes: a Step towards Online Learning~\cite{hu2024training}) and conducted tests using both the ADAM and RLEKF optimizers. For the ADAM optimizer, the learning rate was reduced until it reached $10^{-7}$, at which point training stopped. In contrast, for the RLEKF optimizer, we uniformly trained for 30 epochs. The differing training criteria were set according to the characteristics of each optimizer; notably, the RLEKF optimizer achieved the convergence precision of ADAM within a few epochs. Therefore, to better highlight the differences in training outcomes, we set the training duration to 30 epochs for RLEKF.

From the experimental results, when using the ADAM optimizer, the RMSE of energy across various systems decreased by 6.2\% to 37.5\%, while the RMSE of forces decreased by 0.5\% to 15\%. Training time increased to 185.42\% to 289.75\% compared to fixed activation functions. Both methods had identical learning rate schedules and converged within the same number of epochs, as detailed in the table. When using the RLEKF optimizer, the RMSE of energy decreased by 8.9\% to 39.9\%, and the RMSE of forces decreased by 1.3\% to 5.8\%. Training time increased to 237.10\% to 362.58\% compared to fixed activation functions.
\begin{table}[]
  \begin{center}
    \caption{Convergence Results of the DP Model Using the RLEKF Optimizer on a Large Dataset}
    \label{tab:error_energy}
    \begin{threeparttable}
    \begin{tabular}{lllllllll}\hline
        RLEKF  & ~ & Al & Cu & CuO & Mg & NaCl & Si & H2O  \\ \hline
       ~ & tanh & 2.88E-04 & 2.45E-04 & 1.03E-04 & 1.44E-04 & 1.98E-05 & 1.51E-04 & 9.68E-05  \\ 
        E/atom, eV & Bline & 2.41E-04 & 2.14E-04 & 6.28E-05 & 1.49E-04 & 1.64E-05 & 1.37E-04 & 7.42E-05  \\ 
        ~ & Ratio & 83.74\% & 87.54\% & 60.87\% & 103.71\% & 83.00\% & 91.16\% & 76.69\%  \\ \hline
        ~ & tanh & 3.27E-02 & 4.18E-02 & 2.36E-02 & 1.22E-02 & 3.75E-03 & 2.26E-02 & 1.77E-02  \\ 
        F, eV/$ \mathring{A}$ & Bline & 3.09E-02 & 4.12E-02 & 2.16E-02 & 1.26E-02 & 3.70E-03 & 2.22E-02 & 1.66E-02  \\ 
        ~ & Ratio & 94.23\% & 98.51\% & 91.22\% & 102.83\% & 98.65\% & 98.19\% & 93.89\%  \\ \hline
        ~ & tanh & 384 & 151.7 & 244 & 196 & 1017 & 641.8 & 622  \\ 
        s/epoch & Bline & 712 & 368 & 707 & 399 & 2874 & 1708 & 1614  \\ 
        ~ & Ratio & 185.42\% & 242.58\% & 289.75\% & 203.57\% & 282.60\% & 266.13\% & 259.49\% \\ \hline
    \end{tabular}
    \begin{tablenotes}    
        \item[] Bline denotes thetrainable adaptive activation function using the Bline function. The ratio represents the quotient of the results obtained using thetrainable adaptive activation function compared to those obtained using fixed activation functions.
    \end{tablenotes}         
  \end{threeparttable}
  \end{center}
\end{table}

\begin{table}[]
  \begin{center}
    \caption{Convergence Results of the DP Model Using the ADAM Optimizer on a Large Dataset}
    \label{tab:error_energy}
    \begin{threeparttable}
    \begin{tabular}{lllllllll}\hline
        ADAM  & ~ & Al & Cu & CuO & Mg & NaCl & Si & H2O  \\ \hline
        ~ & tanh & 1.35E-03 & 2.92E-04 & 2.18E-04 & 2.36E-04 & 2.74E-05 & 2.60E-04 & 2.32E-04  \\ 
        E/atom, eV  & Bline & 1.26E-03 & 2.49E-04 & 1.36E-04 & 2.18E-04 & 2.38E-05 & 2.33E-04 & 1.57E-04  \\ 
        ~ & Ratio & 93.82\% & 85.14\% & 62.55\% & 92.48\% & 86.63\% & 89.62\% & 67.67\%  \\ \hline
         ~  & tanh & 3.99E-02 & 4.35E-02 & 2.56E-02 & 1.46E-02 & 3.75E-03 & 2.96E-02 & 2.36E-02  \\ 
        F, eV/$ \mathring{A}$ & Bline & 3.61E-02 & 4.03E-02 & 2.18E-02 & 1.39E-02 & 3.73E-03 & 2.71E-02 & 2.00E-02  \\ 
        ~ & Ratio & 90.65\% & 92.56\% & 85.12\% & 95.47\% & 99.42\% & 91.39\% & 84.84\%  \\ \hline
        ~ & tanh & 172 & 78 & 120 & 99 & 470 & 363 & 296  \\ 
         s/epoch & Bline & 408 & 223 & 420 & 245 & 1704 & 1000 & 804  \\ 
        ~ & Ratio & 237.10\% & 284.65\% & 348.00\% & 246.08\% & 362.58\% & 275.00\% & 271.70\% \\ \hline
    \end{tabular}
  \end{threeparttable}
  \end{center}
\end{table}
Additionally, we visualized the activation function images. It was observed that the activation functions varied across different optimizers within the same system, different systems under the same optimizer, and even between layers. Activation functions evolved continuously during training, sometimes with significant changes. Overall, the magnitude of these changes correlated closely with the convergence trends.

\begin{figure}[t!]
  \centering
  \includegraphics[scale=0.15]{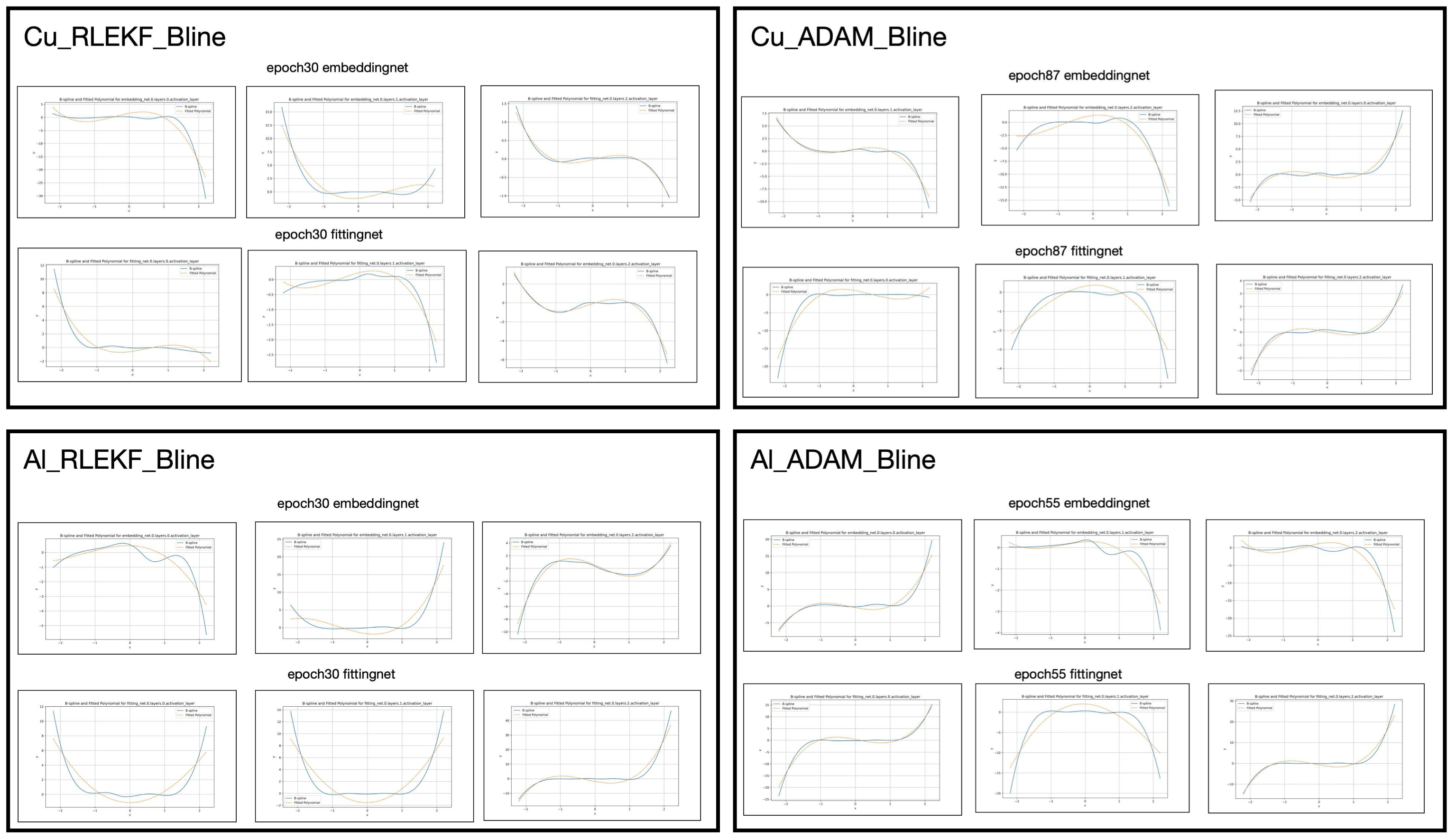}
  \caption{
The solid line represents the activation function fitted using the Bline function, while the dashed line shows the approximation curve of a cubic polynomial.
  }
  \label{fig:bline1}
\end{figure}

\begin{figure}[t!]
  \centering
  \includegraphics[scale=0.135]{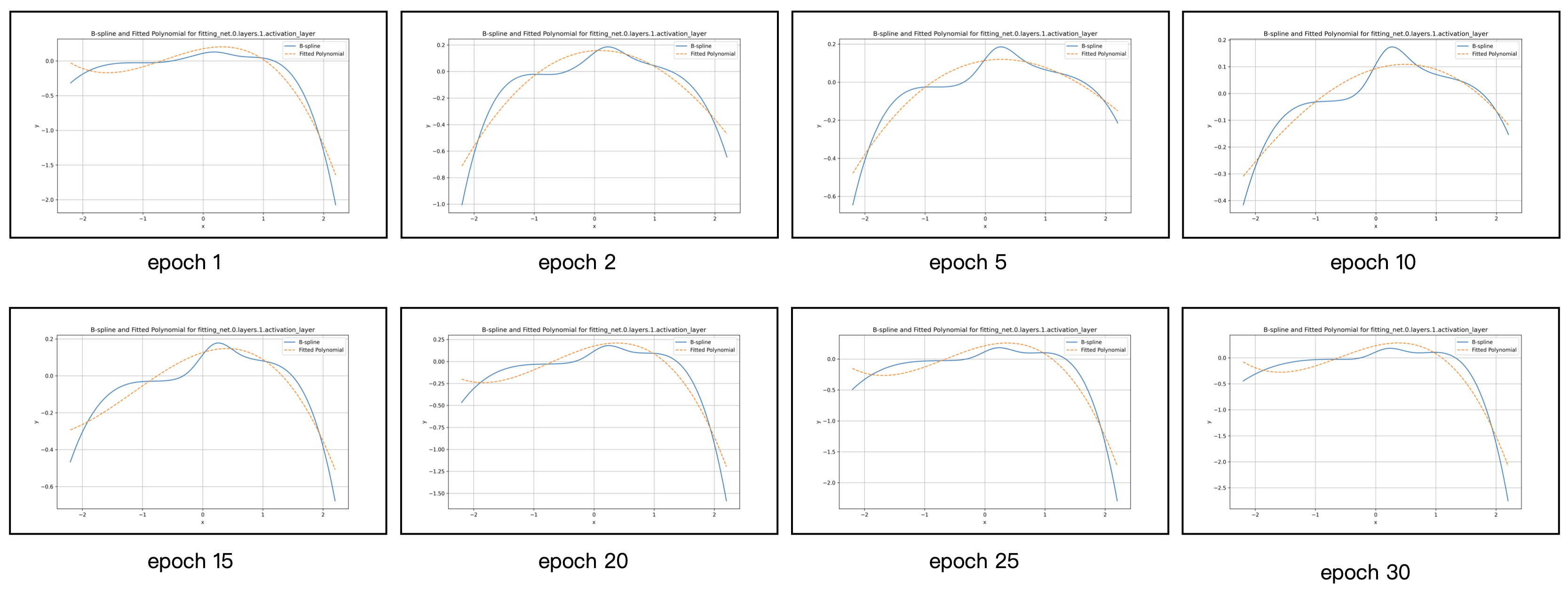}
  \caption{
This depicts the evolution of the activation function for the first layer during fitting. There is a significant change in the second epoch, followed by smaller subsequent variations, which aligns with the convergence trend of the RMSE.
  }
  \label{fig:bline2}
\end{figure}

\subsection{Experiment 3:Mathematical Methods Test}


In Experiment 2, it can be observed that the use of Bline as atrainable adaptive activation function in the DP network significantly increases computational time, approximately doubling or tripling the original duration. This substantial additional overhead is primarily due to the absence of the Bline function in PyTorch, leading to slower computations, further exacerbated by certain characteristics of the function. However, the flexibility offered by Bline is unmatched by other mathematical methods.

In this section, we tested eight mathematical fitting methods, including the Fourier function, using the LKF optimizer for 30 epochs, with copper (Cu) selected as the baseline material. The test results are summarized in the table. Considering comprehensive factors such as energy, force, and computational time, the Fourier and first-kind Chebyshev polynomials performed notably well. Subsequently, we conducted full-system tests as outlined in Experiment 2. The results indicate that both Fourier and first-kind Chebyshev outperformed Bline, with Fourier demonstrating superior performance over Chebyshev. For instance, in CuO, the energy decreased to 38.3 and the force to 78.26 of those obtained using fixed activation functions, while the computational time increased by only about 20.

\begin{table}[]
  \begin{center}
    \caption{Convergence Results for Various Fitting Methods}
    \label{tab:error_energy}
    \begin{threeparttable}
    {\scriptsize
    \begin{tabular}{lllllllllll} \hline
        Cu & Tanh & Bline & Fourier & GRBF & Legendre & Hermite & Chebyshev2 & Chebyshev1 & Bessel & Jacobi  \\ \hline
        E/atom, eV & 2.45E-04 & 2.14E-04 & 1.96E-04 & 2.62E-04 & 2.96E-04 & 2.05E-04 & 2.36E-04 & 2.08E-04 & 2.30E-04 & 2.17E-04  \\ 
        Ratio & 100\% & 87.54\% & 80.17\% & 106.87\% & 120.84\% & 83.89\% & 96.44\% & 84.83\% & 93.85\% & 88.60\%  \\ \hline
          F, eV/$ \mathring{A}$ & 4.18E-02 & 4.12E-02 & 4.18E-02 & 4.42E-02 & 5.23E-02 & 4.12E-02 & 4.27E-02 & 4.12E-02 & 4.25E-02 & 4.17E-02  \\ 
        Ratio & 100\% & 98.51\% & 99.98\% & 105.77\% & 125.12\% & 98.73\% & 102.24\% & 98.73\% & 101.75\% & 99.77\%  \\ \hline
        s/epoch & 151.7 & 368 & 153 & 362 & 139.133 & 183 & 144 & 142 & 179 & 184  \\ 
        Ratio & 100\% & 242.58\% & 100.86\% & 238.63\% & 91.72\% & 120.63\% & 94.92\% & 93.61\% & 118.00\% & 121.29\% \\ \hline
    \end{tabular}
    }
    \begin{tablenotes}    
        \item[] The convergence results for various fitting methods are presented. The DP network was trained using the RLEKF optimizer on the Cu system for 30 epochs.
    \end{tablenotes}         
  \end{threeparttable}
  \end{center}
\end{table}

\begin{table}[]
  \begin{center}
    \caption{Comparison of Convergence Results for Four Fitting Methods Across Multiple Systems}
    \label{tab:error_energy}
    \begin{threeparttable}
    
    \begin{tabular}{lllllllll} \hline
        RLEKF & ~ & Al & Cu & CuO & Mg & NaCl & Si & H2O  \\  \hline
        ~ & Tanh & 2.88E-04 & 2.45E-04 & 1.03E-04 & 1.44E-04 & 1.98E-05 & 1.51E-04 & 9.680E-05  \\ 
        E/atom, eV & Bline & 2.41E-04 & 2.14E-04 & 6.28E-05 & 1.49E-04 & 1.64E-05 & 1.37E-04 & 7.424E-05  \\ 
        ~ & ~ & 83.74\% & 87.54\% & 60.87\% & 103.71\% & 83.00\% & 91.16\% & 76.69\%  \\ 
        ~ & Fourier & 2.20E-04 & 1.96E-04 & 3.95E-05 & 1.12E-04 & 1.31E-05 & 1.29E-04 & 5.418E-05  \\ 
        ~ & ~ & 76.50\% & 80.17\% & 38.30\% & 78.10\% & 66.05\% & 85.56\% & 55.97\%  \\ 
        ~ & CBSV1 & 2.64E-04 & 2.08E-04 & 5.77E-05 & 1.33E-04 & 1.60E-05 & 1.34E-04 & 6.765E-05  \\ 
        ~ & ~ & 91.53\% & 84.83\% & 55.89\% & 92.81\% & 81.14\% & 89.11\% & 69.89\%  \\ \hline
        ~ & Tanh & 3.27E-02 & 4.18E-02 & 2.36E-02 & 1.22E-02 & 3.75E-03 & 2.26E-02 & 1.773E-02  \\ 
        F, eV/$ \mathring{A}$ & Bline & 3.09E-02 & 4.12E-02 & 2.16E-02 & 1.26E-02 & 3.70E-03 & 2.22E-02 & 1.665E-02  \\ 
        ~ & ~ & 94.23\% & 98.51\% & 91.22\% & 102.83\% & 98.65\% & 98.19\% & 93.89\%  \\ 
        ~ & Fourier & 2.98E-02 & 4.18E-02 & 1.85E-02 & 1.15E-02 & 3.67E-03 & 2.21E-02 & 1.506E-02  \\ 
        ~ & ~ & 90.89\% & 99.98\% & 78.26\% & 94.07\% & 97.65\% & 97.81\% & 84.92\%  \\ 
        ~ & CBSV1 & 3.18E-02 & 4.12E-02 & 2.10E-02 & 1.19E-02 & 3.69E-03 & 2.15E-02 & 1.603E-02  \\ 
        ~ & ~ & 97.26\% & 98.73\% & 88.79\% & 97.45\% & 98.42\% & 95.18\% & 90.42\%  \\ \hline
        ~ & Tanh & 384 & 151.7 & 244 & 196 & 1017 & 641.8 & 622  \\ 
        s/epoch & Bline & 712 & 368 & 707 & 399 & 2874 & 1708 & 1614  \\ 
        ~ & ~ & 185.42\% & 242.58\% & 289.75\% & 203.57\% & 282.60\% & 266.13\% & 259.49\%  \\ 
        ~ & Fourier & 395 & 153.6 & 298 & 230 & 1307 & 790 & 879  \\ 
        ~ & ~ & 102.86\% & 101.25\% & 122.13\% & 117.35\% & 128.52\% & 123.09\% & 141.32\%  \\ 
        ~ & CBSV1 & 399 & 142 & 340 & 261 & 1516 & 858 & 889  \\ 
        ~ & ~ & 103.92\% & 93.61\% & 139.34\% & 133.16\% & 149.07\% & 133.69\% & 142.93\% \\ \hline
    \end{tabular}
    
    \begin{tablenotes}    
        \item[] This section compares the convergence results between threetrainable adaptive activation functions and the Tanh function. The percentages indicate the ratios relative to the Tanh results.
    \end{tablenotes}         
  \end{threeparttable}
  \end{center}
\end{table}

We also recorded the number of epochs required to achieve the same energy precision as fixed activation functions and the equivalent epoch times. Except for NaCl, Bline achieved the same precision in less time than fixed activation functions. In contrast, Fourier and first-kind Chebyshev showed even more significant reductions in time compared to fixed methods.

\begin{table}[]
  \begin{center}
    \caption{Comparison of Epochs Required to Reach the Same Convergence Value}
    \label{tab:error_energy}
    \begin{threeparttable}
   
    \begin{tabular}{lllllllll}\hline
        ~ & ~ & Al & Cu & CuO & Mg & NaCl & Si & H2O  \\ \hline
        Tanh & ~ & 30 & 30 & 30 & 30 & 30 & 30 & 30  \\ \hline
        bline & train & 11 & 12 & 10 & $ / $ & 6 & 17 & 11  \\ 
        ~ & Equivalent & 20.40  & 29.11  & 28.98  & $ / $ & 16.96  & 45.24  & 28.54   \\ \hline
        Fourier & train & 6 & 9 & 7 & 10 & 2 & 7 & 5  \\ 
        ~ & Equivalent & 6.17  & 9.11  & 8.55  & 11.73  & 2.57  & 8.62  & 7.07   \\ \hline
        Chebyshev1 & train & 21 & 11 & 11 & 20 & 4 & 12 & 7  \\ 
        ~ & Equivalent & 21.82  & 10.30  & 15.33  & 26.70  & 5.97  & 16.05  & 10.01  \\ \hline
    \end{tabular}
    
    \begin{tablenotes}    
        \item[] Train: The number of epochs required for each method to reach the specified convergence value.

Equivalent: The equivalent number of epochs required to achieve the same convergence value compared to the method using fixed activation functions.
    \end{tablenotes}         
  \end{threeparttable}
  \end{center}
\end{table}

The fitted activation function curves for Fourier and CBSV are illustrated in the figure, where Fourier exhibits more dramatic variations, whereas other methods show smoother curves. Notably, the shapes of the fitted activation function curves differ entirely from those of fixed activation functions. Through Experiments 2 and 3, we have demonstrated that our method employs a data-driven adaptive fitting approach. Compared to othertrainable adaptive activation function methods, it exhibits stronger data dependency and adaptivity.
\begin{figure}[t!]
  \centering
  \includegraphics[scale=0.17]{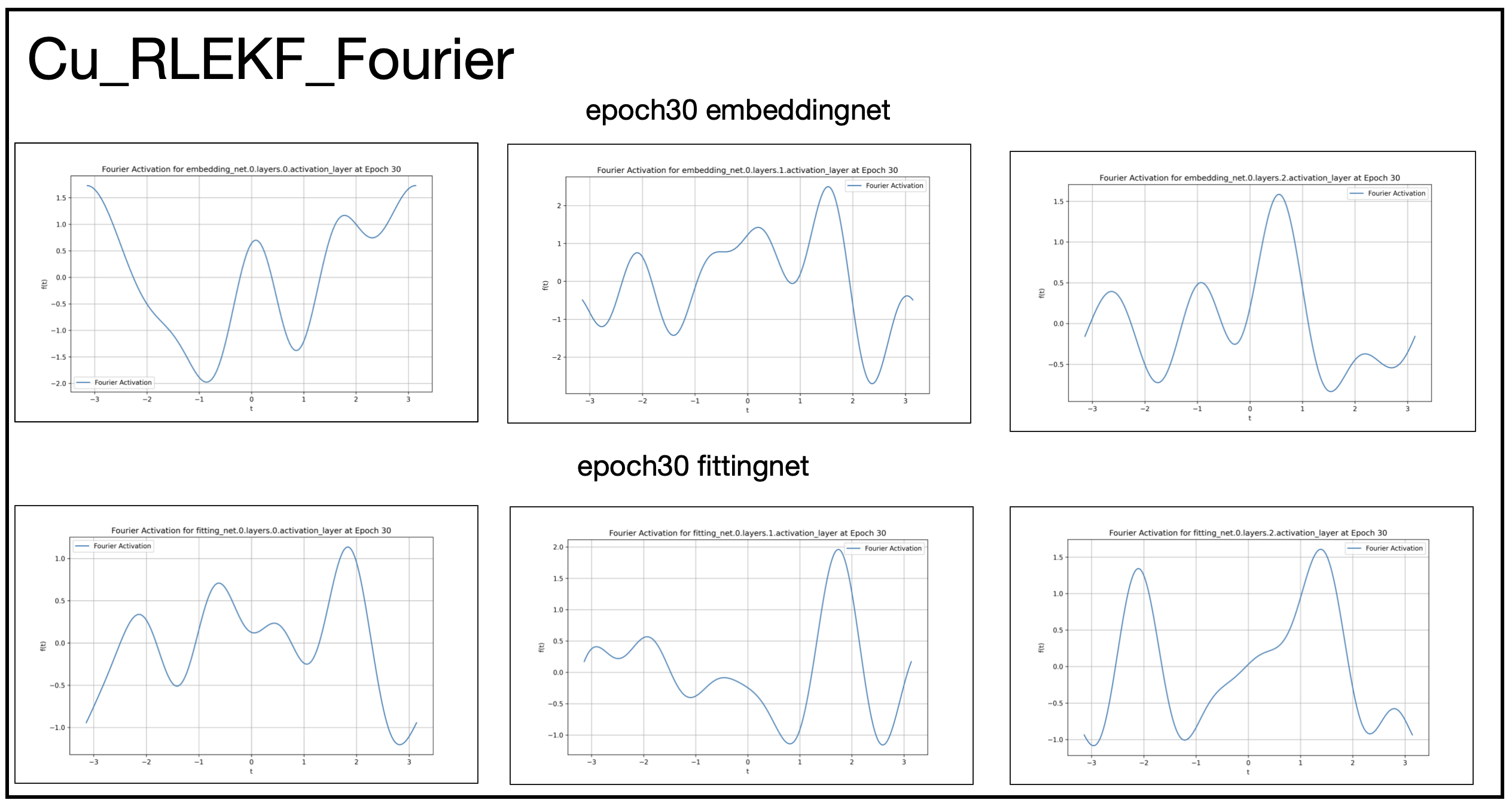}
  \caption{
The activation function fitted using Fourier transform.
  }
  \label{fig:fft}
\end{figure}

\begin{figure}[t!]
  \centering
  \includegraphics[scale=0.17]{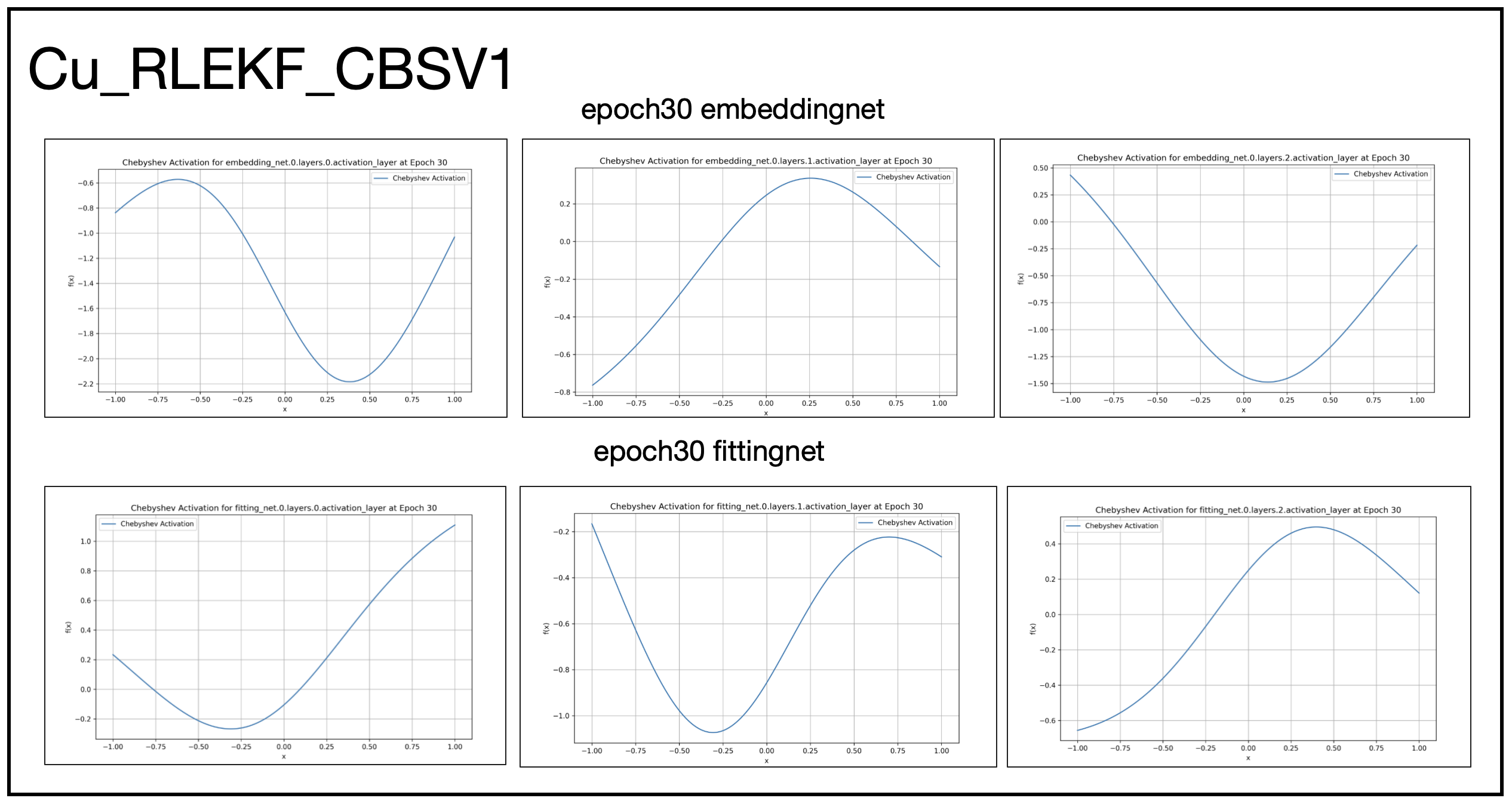}
  \caption{
The activation function fitted using first-kind Chebyshev polynomials.
  }
  \label{fig:CBSV1}
\end{figure}

\subsection{Experiment 4:Molecular dynamics simulations}
To mitigate issues such as overfitting, we utilized LAMMPS for molecular dynamics (MD) simulations. The PWMLFF framework inherently supports a LAMMPS interface, which we employed to conduct MD simulations on four models—Bline, Fourier, CBSV, and Hermite—as detailed in Experiment 3. Each simulation step was set to 1 femtosecond (fs), with a total of 1000 steps executed.

The resulting curves of total energy changes revealed that the Fourier method exhibited significant deviations compared to the other approaches. Additionally, we visualized the molecular trajectories and selected a representative frame for comparison, as shown in the figure. The visualization clearly indicates that the trajectory generated by the Fourier method diverges substantially from those produced by the other methods.

By examining the fitting curves of the activation functions used in the Fourier method, it becomes evident that this type of periodic function, characterized by abrupt changes, is not suitable for tasks within the Deep Potential (DP) framework. This conclusion underscores the importance of selecting appropriate activation functions that align with the specific requirements of the task at hand.

\begin{figure}[t!]
  \centering
  \includegraphics[scale=0.25]{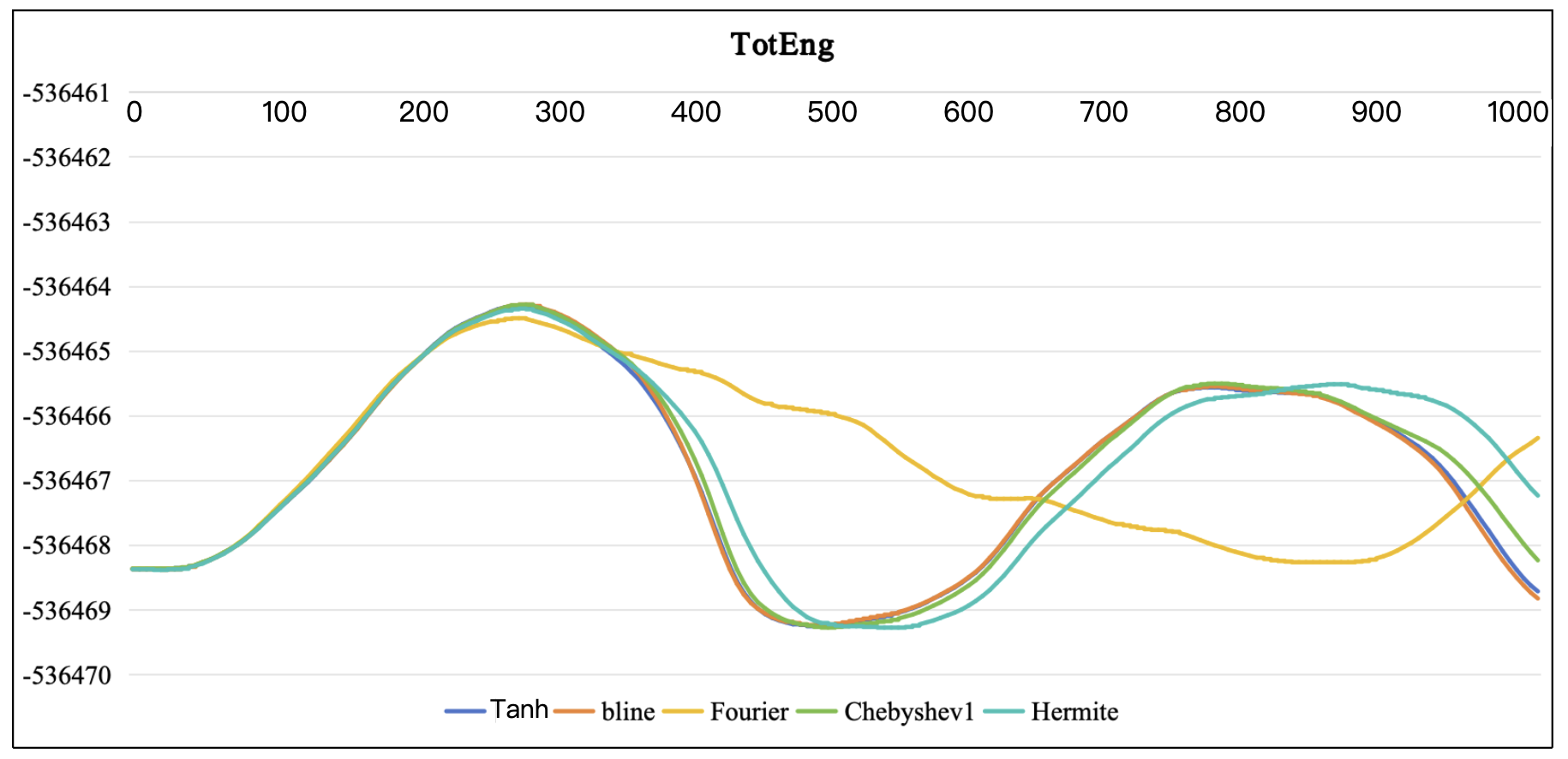}
  \caption{
The total energy variation curves over 1000 steps (with each step being 1 fs) for several models using LAMMPS for molecular dynamics (MD) simulations.
  }
  \label{fig:MD1}
\end{figure}

\begin{figure}[t!]
  \centering
  \includegraphics[scale=0.22]{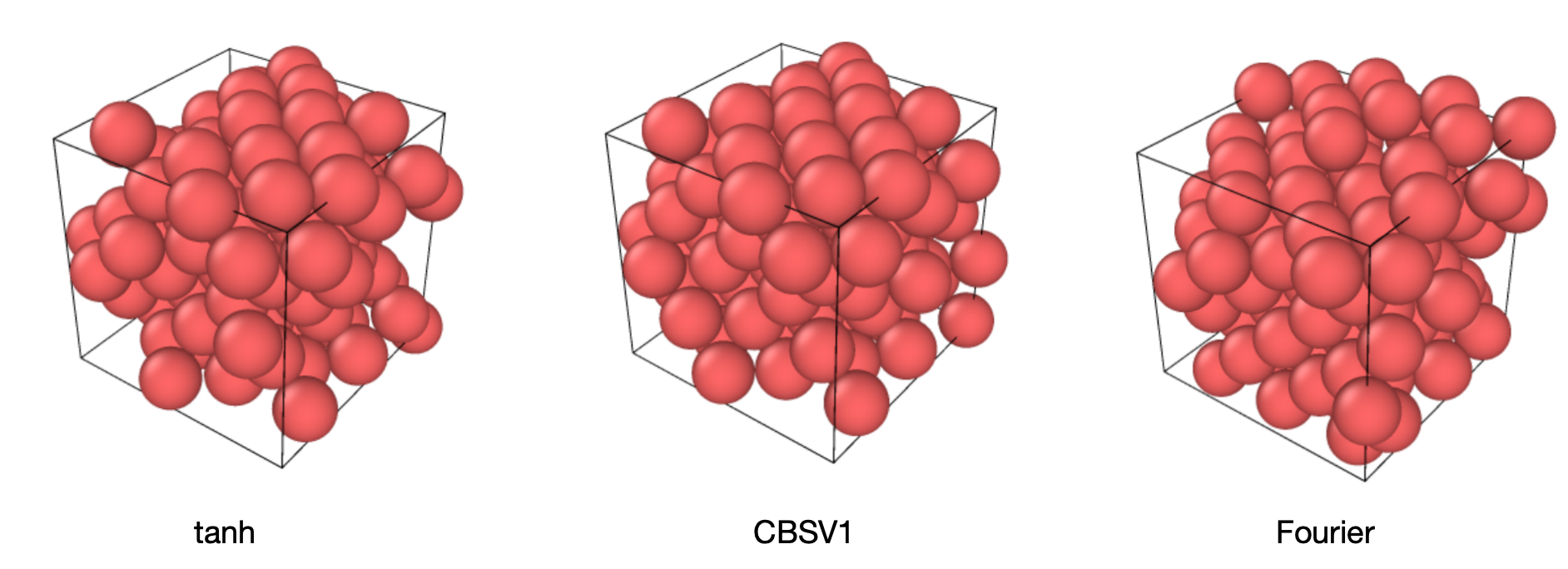}
  \caption{
Visualization of the 957th Frame Trajectory.
  }
  \label{fig:MD2}
\end{figure}

\subsection{Experiment 5:GNN model Test}

The experiments utilized two GNN-based neural network force field models: PAINN and ChgNet. Previous experiments have demonstrated the effectiveness oftrainable adaptive activation functions in standalone MLPs. In the PAINN model, we replaced the SiLU activation function within the updata and message layers withtrainable adaptive activation functions fitted using BLINE and CBSV. Specifically, "L" denotes "num-interaction": "2", which refers to the number of interaction operations; "h" represents "hidden-channels": "128", indicating the number of feature channels in hidden layers (i.e., layers that are neither input nor output); and "r" stands for "num-radial-basis": "16", referring to the number of radial basis functions (RBFs), a class of functions based on inter-point distances.

The results are summarized in the table. From these results, it is clear that the Bline activation function performed best, while CBSV, though inferior to Bline, still outperformed fixed activation functions.
\begin{table}[]
  \begin{center}
    \caption{PAINN Test set results}
    \label{tab:error_energy}
    \begin{threeparttable}
    \tiny
    \begin{tabular}{llll|lll|lll|lll}\hline
         ~ &~ & l2h64r16 & ~ & ~ & l2h64r32 & ~ & ~ & l2h128r16 & ~ & ~ & l2h128r32 &    \\ \hline
        ~ & E\_RMSE & F\_RMSE & Time & E\_RMSE & F\_RMSE & Time & E\_RMSE & F\_RMSE & Time & E\_RMSE & F\_RMSE & Time  \\ 
         Silu & 8.64E-02 & 1.56E-02 & 55712  & 4.78E-02 & 1.44E-02 & 59447  & 7.77E-02 & 1.56E-02 & 59644  & 1.21E-01 & 1.43E-02 & 59838   \\ 
        Bliene & 4.72E-02 & 1.57E-02 & 74177  & 4.85E-02 & 1.49E-02 & 75826  & 5.95E-02 & 1.57E-02 & 79616  & 2.86E-02 & 1.39E-02 & 79422   \\ 
        CBSV1 & 4.85E-02 & 1.60E-02 & 62656  & 4.65E-02 & 1.46E-02 & 62968  & 5.84E-02 & 1.60E-02 & 62995  & 5.17E-02 & 1.43E-02 & 63306   \\ \hline
         ~ &~ & l3h64r16 & ~ & ~ & l3h64r32 & ~ & ~ & l3h128r16 & ~ & ~ & l3h128r32 &    \\ \hline
        ~ & E\_RMSE & F\_RMSE & Time & E\_RMSE & F\_RMSE & Time & E\_RMSE & F\_RMSE & Time & E\_RMSE & F\_RMSE & Time  \\ \hline
         Silu & 6.40E-02 & 1.59E-02 & 65013  & 6.28E-02 & 1.61E-02 & 64813  & 8.80E-02 & 1.58E-02 & 65029  & 4.48E-02 & 1.36E-02 & 65283   \\ 
        Bliene & 4.73E-02 & 1.55E-02 & 94409  & 3.52E-02 & 1.38E-02 & 95403  & 6.25E-02 & 1.50E-02 & 104581  & 3.39E-02 & 1.42E-02 & 104513   \\ 
        CBSV1 & 4.11E-02 & 1.51E-02 & 73434  & 3.33E-02 & 1.37E-02 & 74649  & 8.70E-02 & 1.51E-02 & 71864  & 4.10E-02 & 1.46E-02 & 71252  \\ \hline
    \end{tabular}
            
  \end{threeparttable}
  \end{center}
\end{table}

In the ChgNet model, due to the separate implementation of MLP and GatedMLP functions, where GatedMLP invokes the MLP function, we directly modified the activation functions within the MLP to use Bline and CBSVtrainable adaptive activation functions. Given that this model was pre-trained, using the full MPtr dataset for training would be excessively time-consuming; thus, only the first 1\% of the MPtr data was used for training. The performance improvements were as follows: energy predictions improved by 9\% with Bline and 15\% with CBSV, while force predictions improved by 12\% with Bline and 7\% with CBSV. Regarding parameter counts, the fixed activation function model had 403,126 parameters, whereas the Bline model added only 160 parameters, and the CBSV model added just 64 parameters. Thus, usingtrainable adaptive activation functions in ChgNet led to more than a 10\% performance improvement with an increase of only 0.016\% in parameters.

\begin{table}[]
  \begin{center}
    \caption{chgnet Test set results}
    \label{tab:error_energy}
    \begin{threeparttable}
   
    \begin{tabular}{llllll} \hline
        ~ & e\_MAE & f\_MAE & s\_MAE & m\_MAE & Parameters \\  \hline
        Silu & 0.033 & 0.105 & 0.492 & 0.099 & 403126 \\ 
        Bline & 0.03 & 0.092 & 0.435 & 0.093 & 403286 \\ 
        CBSV1 & 0.028 & 0.087 & 0.479 & 0.087 & 403190 \\ \hline
    \end{tabular}
            
  \end{threeparttable}
  \end{center}
\end{table}

\section{Conclusion}


This paper introduces Trainable Adaptive Activation Function Structure (TAAFS) in neural networks that exhibits adaptivity through polynomial fitting of data. This approach has demonstrated promising results in networks containing multiple multi-layer perceptrons (MLPs), such as Deep Potential (DP) and ANI-2, while also showing improvements in models like PAINN and ChgNet. Overall, this method effectively enhances the representation capability of neural networks with only a marginal increase in parameters. However, the effectiveness of the enhancement is data-dependent, and different fitting strategies can influence task performance. Consequently, it is recommended to select fitting methods based on the characteristics of the specific task at hand. We have also explored adjusting parameters of the TAF, such as grid numbers or iteration counts, which impact the training outcomes. Nonetheless, in our experiments, we primarily used the most common parameter values without extensive tuning. Additionally, molecular dynamics (MD) simulations conducted within the DP framework confirmed the generalization ability of the proposed method.

We welcome contributions to expand the repertoire of mathematical formulas. We also welcome the use of this approach in various types of tasks, with a focus on summarizing the compatibility between different mathematical formulas and specific tasks.



\bibliographystyle{unsrtnat}
\bibliography{ref}

\end{document}